\documentclass[hidelinks, 11pt]{article}
\usepackage[margin=1in]{geometry}
\usepackage[utf8]{inputenc}
\usepackage{authblk}

\usepackage{setspace}
\usepackage{amsmath}
\usepackage{amssymb} 
\usepackage{algorithm}
\usepackage{algorithmic}
\usepackage{comment}
\usepackage{setspace}

\usepackage{graphicx}
\usepackage{caption}
\usepackage{subcaption}
\usepackage{multirow}
\usepackage{adjustbox}
\usepackage[hyphens]{url}
\usepackage{hyperref}
\hypersetup{
  colorlinks   = true,
  urlcolor     = red,
  linkcolor    = blue, 
  citecolor   = blue 
}
\usepackage[numbers]{natbib}
\floatstyle{plain}
\allowdisplaybreaks

\def\argmax{{\arg\max}}

\def\bSigma{{\boldsymbol\Sigma}}

\def\ba{\mathbf{a}}

\def\bu{\mathbf{u}}
\def\bv{\mathbf{v}}
\def\bw{\mathbf{w}}
\def\bx{\mathbf{x}}
\def\by{\mathbf{y}}
\def\bz{\mathbf{z}}

\def\btheta{{\boldsymbol\theta}}

\def\bmu{{\boldsymbol\mu}}

\def\bxi{{\boldsymbol\xi}}

\def\bsigma{{\boldsymbol\sigma}}

\def\bphi{{\boldsymbol\phi}}

\def\bbE{\mathbb{E}}

\def\bbH{\mathbb{H}}

\def\bbR{\mathbb{R}}

\def\cC{\mathcal{C}}
\def\cD{\mathcal{D}}

\def\cJ{\mathcal{J}}

\def\cL{\mathcal{L}}

\def\cN{\mathcal{N}}

\def\cP{\mathcal{P}}

\def\cT{\mathcal{T}}

\def\cat{{\sf cat}}

\def\diag{{\sf diag}}

\def\eps{{\epsilon}}

\def\KL{{\sf KL}}

\def\unif{{\sf unif}}

\def\var{{\sf var}}

\def\model{\texttt{LANP}}

\def\lnp{\texttt{LNP}}
\def\anp{\texttt{ANP}}
\def\mgp{\texttt{VMGP}}
\def\spacingset#1{\renewcommand{\baselinestretch}%
{#1}\small\normalsize} \spacingset{1}

\title{Real-time Adaptation for Condition Monitoring Signal Prediction using Label-aware Neural Processes}

\author[1]{Seokhyun Chung\thanks{Corresponding author: schung@virginia.edu}}

\affil[1]{Department of Systems \& Information Engineering, University of Virginia}

\author[2]{Raed Al Kontar}

\affil[2]{Department of Industrial \& Operations Engineering, University of Michigan}

\date{}

\begin{document}
\sloppy
\maketitle

\begin{abstract}
Building a predictive model that rapidly adapts to real-time condition monitoring (CM) signals is critical for engineering systems/units. Unfortunately, many current methods suffer from a trade-off between representation power and agility in online settings. 
For instance, parametric methods that assume an underlying functional form for CM signals facilitate efficient online prediction updates. However, this simplification leads to vulnerability to model specifications and an inability to capture complex signals. On the other hand, approaches based on over-parameterized or non-parametric models can excel at explaining complex nonlinear signals, but real-time updates for such models pose a challenging task. 
In this paper, we propose a neural process-based approach that addresses this trade-off. It encodes available observations within a CM signal into a representation space and then reconstructs the signal's history and evolution for prediction. Once trained, the model can encode an arbitrary number of observations without requiring retraining, enabling on-the-spot real-time predictions along with quantified uncertainty and can be readily updated as more online data is gathered. Furthermore, our model is designed to incorporate qualitative information (i.e., labels) from individual units. This integration not only enhances individualized predictions for each unit but also enables joint inference for both signals and their associated labels. 
Numerical studies on both synthetic and real-world data in reliability engineering highlight the advantageous features of our model in real-time adaptation, enhanced signal prediction with uncertainty quantification, and joint prediction for labels and signals.
\end{abstract}

Keywords: real-time adaptation; neural processes; meta-learning; condition monitoring; degradation

\newpage
\spacingset{1.5} 
\section{Introduction}

Data-driven predictive modeling of condition monitoring (CM) signals is a key component of reliability analysis for engineering systems/units nowadays. It involves building a model that encodes knowledge from historical CM signals and then utilizing it to predict the future evolution of CM signals from an in-service unit with \textit{online} observations collected up until the current point. In this framework, refining predictions to accommodate online observations is instrumental in capturing unit-specific patterns inherent in the CM signal. Such an online update process is often referred to as ``adaptation'' or ``personalization'' since it aims to tailor predictions to each individual unit under consideration, given its real-time data collected.  

However, advances in modern systems pose key challenges in adaptation. CM data collected through Internet of Things (IoT) sensors is often acquired in real-time and at a high frequency, necessitating immediate adaptation to online observations. At the same time, the ever-increasing complexity of engineering systems gives rise to complicated and nonstationary CM signals, demanding models with strong representational capabilities. Unfortunately, existing methods suffer from a trade-off between representation power and agility to online data. Methods that assume a simplified form for CM signals, such as polynomial mixed-effects models \cite{gebraeel2005residual}, can perform efficient online updates but are susceptible to model misspecification and struggle when modeling highly non-linear signals. On the other hand, methods with strong representational power, such as over-parameterized \cite{zhang2018long} or non-parametric models \cite{kontar2017nonparametric}, are hard to adapt in real-time and often require retraining the predictive model when new data is collected. As a consequence, their broad application is greatly constrained in recent CM practices that demand both agility and rich expressiveness, \textit{simultaneously}.


Another significant challenge in adaptation emerges when units are of different types or conditions, identified by categorical information (i.e., labels). To illustrate, consider the CM of electric vehicle batteries. These batteries may be from different production lots or original equipment manufacturers, or they may operate within vehicles of different brands, models, or generations. The presence of such categorical disparities often results in significant variations in CM signals among units based on their respective labels. That said, in practices such as degradation-based CM, these variations remain dormant initially and become evident only at a later stage due to cumulative effects \cite{chung2020functional}. For example, while initially presenting as healthy, lithium-ion batteries from a disqualified lot often exhibit abnormal degradation trends at the later stage, unacceptably deviating from the degradation trend of normal batteries, as presented in our case study in Sec. \ref{sec:model_val}. In such cases, an early-stage adaptation that relies solely on indistinguishable initial trends can easily fail to learn future deviations. This necessitates leveraging label information from the units along with their CM signals. 


This article aims to address both challenges in adaptation outlined above -- real-time personalization for complex signals and the incorporation of label information. 
Specifically, we present a real-time prognostic approach that \textit{dynamically predicts and quantifies uncertainty in real-time} as data is collected from an in-service unit. The real-time adaptivity comes with augmented predictive capabilities by incorporating label information. In the spirit of representation learning, our approach learns an encoder that maps a CM signal along with its label into a representation space, and a decoder that reconstructs the signal using the encoded representation enriched by label information. Notably, the encoder is trained to efficiently encode signals containing any quantity of contextual observations. That is, a trained encoder can encode an arbitrary number of online observations. This, in turn, facilitates an instantaneous update of predictions upon the arrival of new data without the need for any retraining procedures. This advantage is underpinned by the strong representation power achieved by using deep neural networks (DNNs) to parameterize the encoders and the decoder. Our framework does not impose a requirement for all units to possess labels; it remains applicable even when some units have missing label information. Our label-aware modeling not only significantly improves personalized predictions but also facilitates the joint prediction of the unit's future signal evolution and its label when missing. We also offer a data augmentation approach for cases where the number of historical units is insufficient for training our model.

We summarize our contributions as follows.

\begin{itemize}
    \item We propose an NP-based framework for predictive modeling of CM data. Our model is highly scalable and achieves instantaneous predictions and uncertainty quantification in real-time as data is collected from an in-service unit.
    
    \item We propose label-aware modeling to seamlessly integrate label information from units into our prognostic framework while allowing for missing label information. The integration is shown to substantially enhance predictive power. Notably, our framework can jointly predict future CM signal evolution and a unit's label when it is missing. 
        
    
    \item  We validate our framework using synthetic data and a real-world reliability engineering dataset. Results demonstrate the advantageous features of our model in fast adaptation, enhanced predictions, and uncertainty quantification. In particular, the case study underscores the appealing capability of our model to make joint predictions for curves and labels. It also highlights the ability to make accurate predictions in the early stages, which may be critical for early anomaly detection within in-service units. 
\end{itemize}

The remainder of the paper is organized as follows. Sec. \ref{sec:anp} sets the stage by introducing NPs within our problem context. Sec. \ref{sec:prop_model} introduces the proposed approach. Sec. \ref{sec:related work} discusses relevant literature. Sec. \ref{sec:model_val} assesses the proposed model using simulation and real-world data. 
Sec. \ref{sec:conclusion} concludes the paper.

\section{Background: NPs}\label{sec:anp}

The NP is a group of models that learn a distribution over functions \cite{jha2022neural}. Given a set of functions as a training set, the training process of NPs can be viewed as learning a prior distribution over the functions. Then in the prediction stage, it estimates a specific function by deriving a posterior predictive distribution given its available observations. In the context of CM, the training process learns a distribution over historical CM signals, and the prediction process estimates the CM signal of the in-service unit based on its online observations. Before delving into our proposed model, this section situates NPs within a general framework, where we introduce how NPs construct a probabilistic model for a single arbitrary function. 

\subsection{Modeling a function using NPs} \label{sec:NPs}

Consider an \textit{arbitrary} regression function $f:\bbR^{d_{\sf in}} \rightarrow \bbR^{d_{\sf out}}$ sampled from a set of functions that share the same domain. Let $\cD_i = (\bx_i, \by_i)$ denote an input-output pair, indexed by $i$, where $\bx_i \in \bbR^{d_{\sf in}}$ and output $\by_i \in \bbR^{d_{\sf out}}$ are a respective input-output pair from the function. We will call such a tuple an observation. The collection of available observations from the function is called \textit{contexts} $\cD_\cC:= \{\cD_i\}_{i\in \cC}$. The input-output pairs for which predictions are to be made are called \textit{targets} $\cD_\cT:=\{\cD_i\}_{i\in \cT}$; here $\cC$ and $\cT$ denote the index sets of contexts and targets, respectively. For conciseness, we further define collective notations $\bx_\cC := \{\bx_i\}_{i\in \cC}$, $\bx_\cT := \{\bx_i\}_{i\in \cT}$, $\by_\cC := \{\by_i\}_{i\in \cC}$, $\by_\cT := \{\by_i\}_{i\in \cT}$, and $\cD := \cD_\cC \cup \cD_\cT$. 

The first member of the NP family was proposed by \cite{garnelo2018conditional}. Motivated by the fact that the true trajectory of a regression function should be independent of the order of its context points, they model the targets as a conditional distribution given the contexts while being invariant to the contexts' order. Specifically, the NP builds a distribution conditioned on $\bu_{\cC} \in \bbR^{d_u}$, a finite-dimensional representation of the contexts $\cD_\cC$ invariant to the permutation in $\cC$, that is, $p(\by_\cT \vert \bx_\cT, \cD_\cC) := p(\by_\cT \vert \bx_\cT, \bu_{\cC})$. This is enabled by introducing a deep neural network $u(\cdot)$ that maps each context point to a finite-dimensional representation $u(\cD_i) = \bu_i$ after which a permutation invariant operation is performed over the representations. In particular, the NP employs averaging $\bu_{\cC} = \frac{1}{\vert \cC \vert} \sum_{i\in \cC}\bu_i$ and thus, $\bu_{\cC}$ is invariant to the permutation of $\cD_\cC$. The likelihood $p(\by_\cT \vert \bx_\cT, \bu_{\cC})$ can be formulated as a Gaussian distribution factorized over the target points, represented as 
\begin{align*}
    p(\by_\cT \vert \bx_\cT, \cD_\cC) := p(\by_\cT \vert \bx_\cT, \bu_{\cC}) = \prod_{i\in\cT}\cN(\by_i ; \bmu_{\sf d}, \bsigma_{\sf d}^2) = \prod_{i\in\cT}\cN(\by_i ; \mu_{\sf d}(\bx_i, \bu_{\cC}), \diag(\sigma_{\sf d}^2(\bx_i, \bu_{\cC}))),
\end{align*}
such that mean $\bmu_{\sf d}$ and variance $\bsigma_{\sf d}^2$ functions are parameterized by respective DNNs $\mu_{\sf d}(\cdot)$ and $\sigma^2_{\sf d}(\cdot)$, where 
each outputs a vector in $\bbR^{d_{\sf out}}$. The notation $\diag(\ba)$ indicates the diagonal matrix corresponding to the vector $\ba$.


{
}





\begin{figure}[h!]
 \centering
     \includegraphics[width=0.8\columnwidth]{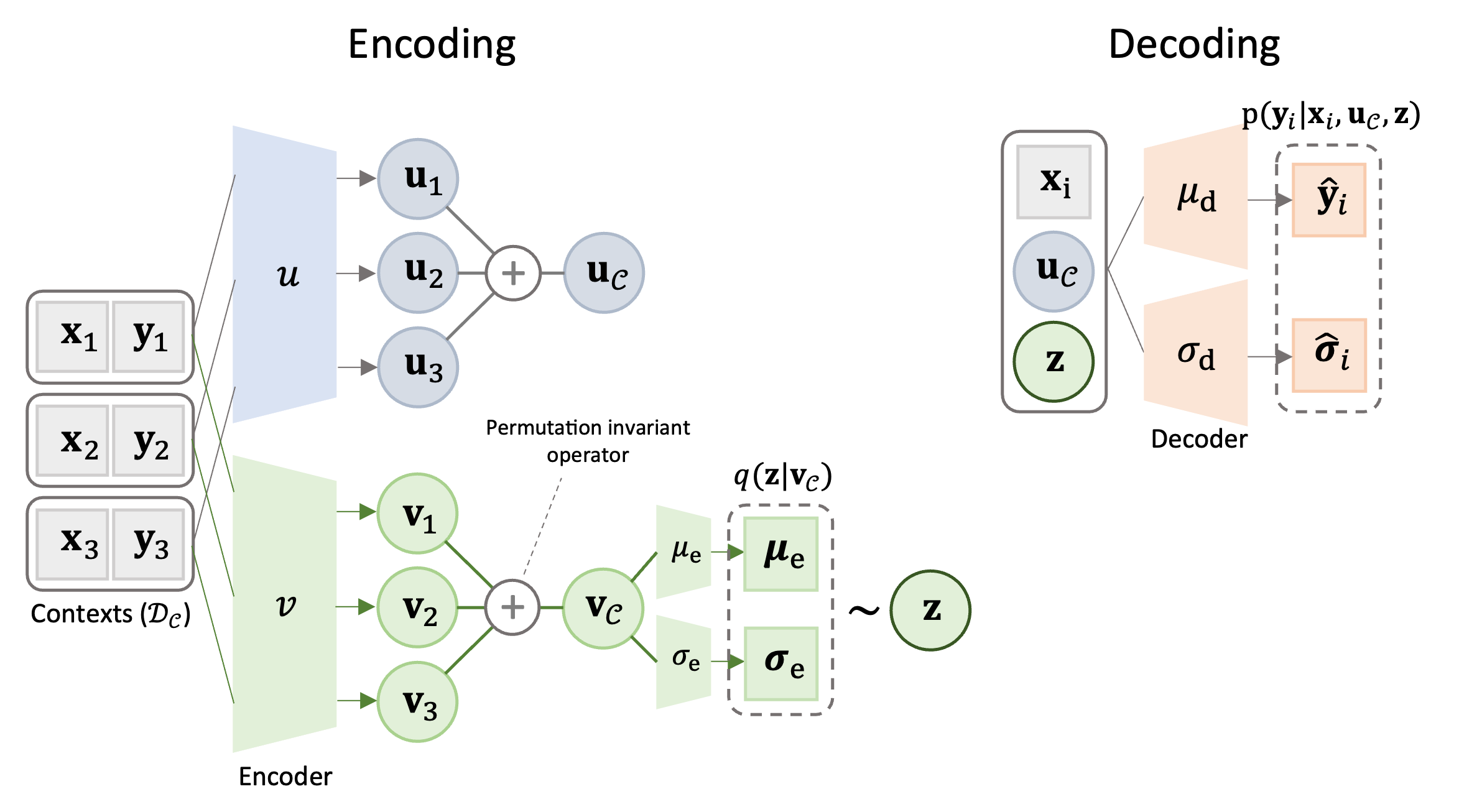}
 \caption{A schematic illustration of \lnp{}.
 }\label{fig:lnp}
 \end{figure}

Notable subsequent work by \cite{garnelo2018neural} introduces a finite-dimensional global latent variable $\bz \in \bbR^{d_{\sf lat}}$ characterized by a factorized Gaussian. This latent variable version of NPs is often called the latent NP (\lnp{}). By incorporating $\bz$, \lnp{} can consider the potential uncertainty in mapping $\cD_\cC$ to the latent space. 
It models the likelihood as being conditioned on $\bz$, and then derives the marginal likelihood by marginalizing $\bz$ out. Specifically, the likelihood conditioning on $\bz$ is expressed as 
\begin{align}\label{eq:p_y_z}
   &p(\by_\cT\vert \bx_\cT, \cD_\cC,\bz) = p(\by_\cT\vert \bx_\cT, \bu_{\cC},\bz) = \prod_{i\in \cT}\cN(\by_i; \mu_{\sf d} (\bx_i, \bu_{\cC}, \bz), \diag(\sigma^2_{\sf d}(\bx_i, \bu_{\cC}, \bz))).
\end{align}
Unlike common latent variable models that involve a prior distribution of $\bz$ in the marginalization, \lnp{}s employ a distribution of $\bz$ conditioned on $\cD_\cC$. The rationale is to introduce a mapping of the contexts $\cD_\cC$ to the latent representation space in a way to be permutation-invariant to $\cC$. The conditional distribution is written by \begin{equation}\label{eq:q_z}
    q(\bz\vert \cD_\cC) := q(\bz \vert \bv_{\cC}) 
    = \cN(\bz; \mu_{\sf e}(\bv_{\cC}), \diag(\sigma^2_{\sf e}(\bv_{\cC})))
\end{equation} where $\bv_{\cC} = \frac{1}{\vert \cC \vert} \sum_{i\in \cC}\bv_i$ is obtained in a similar way to $\bu_\cC$ yet using a DNN $v(\cdot)$, after which $\bv_{\cC}$ is fed to the DNNs $\mu_{\sf e}(\cdot)$ and $\sigma^2_{\sf e}(\cdot)$ that parameterize the mean and variance of the factorized Gaussian $q(\bz\vert \cD_\cC)$, respectively. This results in that $\bv_\cC$ is invariant to the permutation in $\cD_\cC$ and so is $q(\bz\vert \cD_\cC) := q(\bz\vert \bv_\cC)$. 
Given $q(\bz \vert \bv_{\cC})$, the marginal likelihood modeled by \lnp{} is obtained by
\begin{equation}\label{eq:mrg_likl}
    p(\by_\cT \vert \bx_\cT, \cD_\cC) := \int p(\by_\cT\vert \bx_\cT, \bu_{\cC},\bz)q(\bz \vert \bv_{\cC})d\bz.
\end{equation}
In NP literature, $u(\cdot)$, $v(\cdot)$ and $q(\bz|\cdot)$ that encode contexts into a representation space are referred to as the \textit{encoders}, while $p(\by_\cT|\bx_\cT, \cdot)$ that reconstructs a function from a representation is referred to as the \textit{decoder}. Fig. \ref{fig:lnp} illustrates an \lnp{} model with the latent variable $\bz$. 

So far, we have seen how a single arbitrary signal is modeled within the NP framework. In the next section, we will discuss our NP-based method that integratively models a CM signal along with its label, and then see its estimation on a set of historical signals and labels. 




\section{Proposed Approach}\label{sec:prop_model}

Now we introduce \model{}, our proposed model. In Sec. \ref{sec:lanp}, we establish  \model{} built upon hierarchical modeling and derive a variational lower bound for model inference using amortized VI. 
In Sec. \ref{sec:prac}, we discuss some practical considerations in the implementation of \model{}. 




\subsection{\model{}: Label-aware Neural Processes}\label{sec:lanp}

Consider a set of historical units/systems with collected CM signals from each. Suppose that units are categorized into classes where the categorization is inherently reflected in the evolution of their CM signals. For instance, units can be tested as abnormal or normal, with their CM trends exhibiting some heterogeneity according to their labels (i.e., normal vs. abnormal). We consider a case such that label information is available only for some units. Our proposed model, \model{}, trains on available data for historical units in a way that incorporates the possibly incomplete labels into the NP framework to facilitate joint learning of distributions for both signals and labels. This results in the enhancement of signal predictions as well as the capability for label estimation. After training \model{} on historical data, \model{} can perform real-time joint inference for (i) future CM signal evolution and (ii) the label of an in-service unit, given its online observations as contexts.

\subsubsection{Notations}
We start with defining notations. Consider units/systems indexed by $j \in \cJ_{\sf all} := \cJ \cup \{r\}$ that comprises the index set for historical units $\cJ = \{1,..., J\}$ and an in-service unit $\{r\} \not\subset \cJ$. While each unit collects its own CM signal, the full CM signal is available for the historical units $j\in \cJ$, yet an online CM signal collected only until the present time is available for the in-service unit $r$. Now, let us focus on the CM signals of historical units $j\in \cJ$. We denote the index set of contexts and targets among signal observations of unit $j$ as $\cC^j$ and $\cT^j$, respectively. Without loss of generality, we randomly split all available CM observations $\cD^{j}$ into contexts and targets for each historical unit. That is, we denote the targets by $\cD^{j}_\cT := \{\cD^{j}_i\}_{i\in \cT^j} = \{(\bx^{j}_i, \by^j_i)\}_{i \in \cT^j}$ and the context by $\cD^j_\cC := \{\cD^{j}_i\}_{i\in \cC^j} = \{(\bx^{j}_i, \by^j_i)\}_{i \in \cC^j}$ where $\cD^{j} = \cD^j_\cC \cup \cD^j_\cT$.
While for now we generally regard contexts as any random subset, we will shortly discuss a practical selection strategy in the training stage in Sec. \ref{sec:prac}. 

In addition to CM signals, suppose class labels $c^j \in \{1,...,L\}$ are available for some historical units $j\in\cJ_{\sf L} \subset \cJ$ whereas unavailable for units $j\in\cJ_{\sf U} \subset \cJ$. Note that $\cJ_{\sf L} \cup \cJ_{\sf U} = \cJ$ and $\cJ_{\sf L} \cap \cJ_{\sf U} = \varnothing$. Encompassing both labels and CM signals, we can collectively define all available historical data as $\cD^{\sf \cJ} := \left(\{c^j\}_{j\in \cJ_{\sf L}}, \{\cD^j\}_{j\in \cJ}\right)$. Here, what we are interested in, is building an NP model that exploits historical data $\cD^{\sf \cJ}$ to predict the future CM signal trajectory of the in-service unit $r$ as well as its label, given its real-time observations collected as contexts $\cD^r_\cC$.

Recall that in CM, $\bx^{j}_i$ often denotes the $i$-th observational time point at which a continuous signal $\by^j_i$  (such as degradation) from unit $j$ is recorded. Nevertheless, it's important to note that $ \bx^{j}_i \in \bbR^{d_{\sf in}} $ need not solely represent temporal aspects; it can also incorporate various other inputs. Additionally, 
our methodology does not presuppose uniform data collection time points across units, nor does it require units to share the same number of observations. 










\subsubsection{Model development}

Now let us build \model{}. Our goal is to learn an NP model capable of mapping the target points $\cD^j_\cT$ by leveraging the available contexts $\cD^j_\cC$ as well as label information $c^j$ if available\footnote{The notation $c^j$ is abused to denote its corresponding one-hot encoding when it is used as an input to a DNN.}. We start by discussing modeling an arbitrary single historical unit that is with or without label information. Then, we will build an integrated model that encompasses all historical units. For notational brevity, we will temporarily drop $j$ from the notations when discussing modeling for a CM signal from a single arbitrary unit $j$ with or without label information.

\paragraph{Modeling a single CM signal with label information.} Let us first consider an arbitrary historical unit with label information.   
Specifically, we consider a likelihood that includes a latent variable $\bz$ written as
\begin{align}\label{eq:q_y}
  &p(\by_\cT \vert \bx_\cT, c, \bu_\cC,  \bz)  := \prod_{i\in\cT}\cN(\by_i ; \mu_{\sf d}(\bx_i, c, \bu_\cC, \bz), \sigma^2_{\sf d}(\bx_i, c, \bu_\cC, \bz)),  
\end{align}
conditioning on both $\cD_\cC$ and $c$. 
We then introduce a conditional distribution for $\bz$ and $c$ by conditioning on $\cD_\cC$ as follows: 
\begin{align}
    q(c \vert \cD_\cC) &:= q(c \vert \bw_\cC) = \cat(c; \phi(\bw_\cC))\label{eq:q_c}\\
    q(\bz \vert c, \cD_\cC) &:= q(\bz \vert c, \bv_\cC) = \cN(\bz; \mu_{\sf e}(c, \bv_\cC), \diag(\sigma^2_{\sf e}(\bv_\cC)))\label{eq:q_z} 
\end{align}
where $\bw_\cC = \frac{1}{\vert\cC\vert}\sum_{i\in \cC} w(\cD_i)$ is a finite-dimensional representation obtained similarly to $\bu_\cC = \frac{1}{\vert C \vert} \sum_{i\in \cC} u(\cD_i)$ and $\bv_\cC = \frac{1}{\vert C \vert} \sum_{i\in \cC} v(\cD_i)$ but with a DNN $w(\cdot)$ and thus, invariant to the permutation of $\cD_\cC$; $\phi(\cdot)$ indicates a DNN that outputs a probability parameter vector of the categorical distribution with $L$ classes.



The underlying intuition of introducing the above conditional distributions is as follows. Eq. \eqref{eq:q_c} implies that estimating the DNNs $w(\cdot)$ and $\phi(\cdot)$ in $q(c \vert \cD_\cC)$ allows for inferring a unit's label based on its available CM observations $\cD_\cC$. Eq. \eqref{eq:q_z} presents that the distribution of $\bz$ is inferred based on both label information and available CM observations of the unit. In particular, the mean depends on both $c$ and $\cD_\cC$, which encourages the distributions for $\bz$ of different units to be clustered according to their labels. This is sensible in that both $c$ and $\cD_\cC$ would be needed to model the class-level heterogeneity of CM signals. While, the variance depends on $\cD_\cC$ only as $\cD_\cC$ would suffice to infer unit-level deviations within the class.

\begin{figure}[h!]
 \centering
     \includegraphics[width=.8\columnwidth]{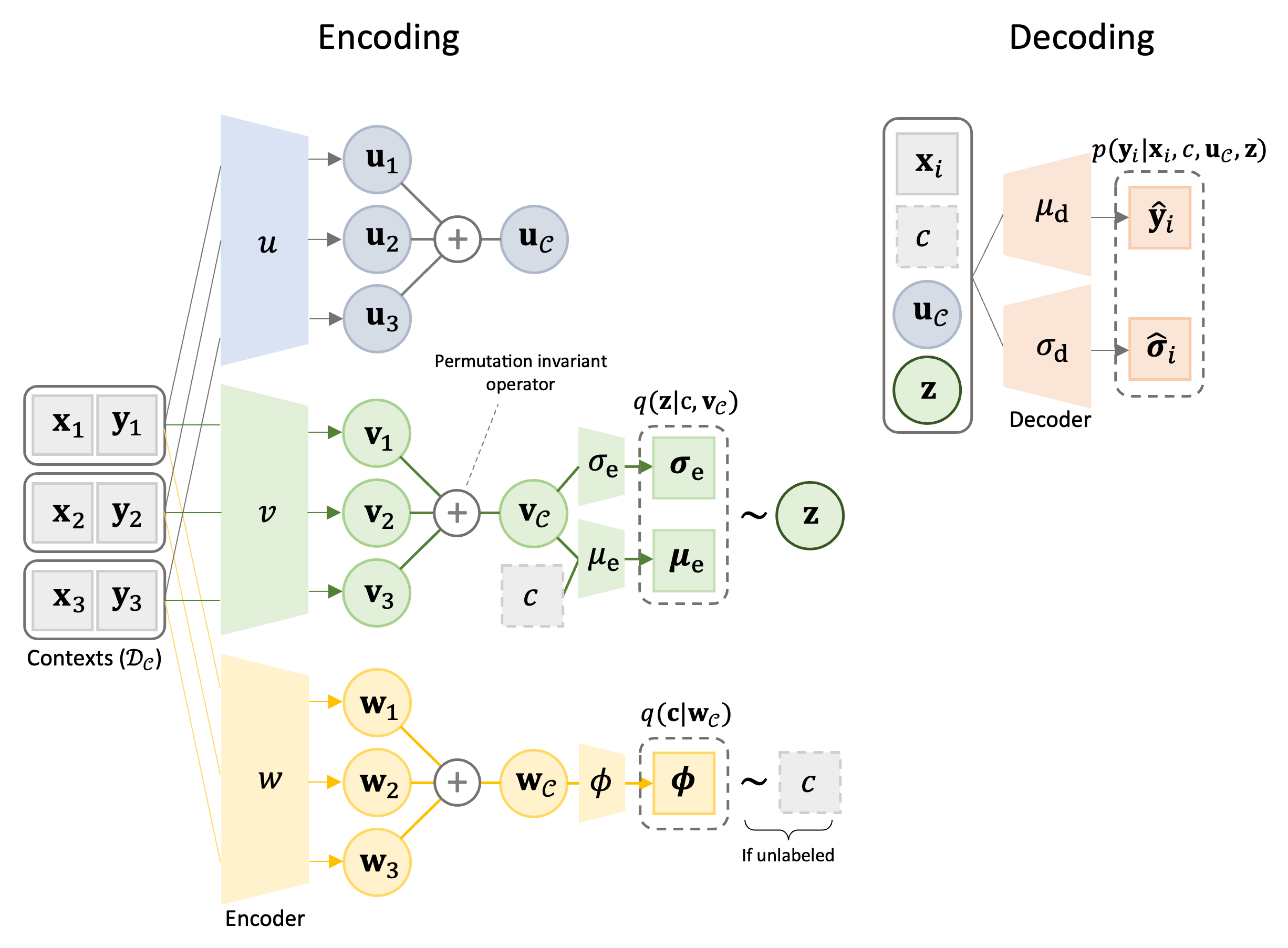}
 \caption{A schematic illustration of \model{}.
 }\label{fig:lanp}
 \end{figure}

Given \eqref{eq:q_y}, \eqref{eq:q_c}, and \eqref{eq:q_z}, the marginal likelihood is given by 
\begin{align}\label{eq:p_y_lab}
    p(\by_\cT, c \vert \bx_\cT, \cD_\cC) := q(c \vert \bw_\cC) \int p(\by_\cT \vert \bx_\cT, c, \bu_\cC, \bz) q(\bz \vert c, \bv_\cC) d\bz.
\end{align}
Fig. \ref{fig:lanp} illustrates the encoding and decoding process of \model{}.

Parameterizing the likelihood by DNNs renders the marginalization \eqref{eq:p_y_lab} for maximum likelihood intractable. Instead, we resort to amortized variational inference (VI) \cite{zhang2018advances}. The framework maximizes a lower bound of the marginal log-likelihood, referred to as the evidence lower bound (ELBO), as an alternative to the original marginal log-likelihood. We refer the reader to \cite{zhang2018advances} for details of VI. The derivation of the ELBO builds upon Jensen's inequality, written as  
\begin{align}
 &\log p(\by_\cT, c \vert \bx_\cT, \cD_\cC) \nonumber \\ 
 & \;\;\;\;\; = \log \int \frac{p(\by_\cT \vert \bx_\cT,c,\bu_\cC, \bz) q(\bz \vert c, \bv_\cC) }{ q(\bz\vert c, \bv_\cT)} q(\bz\vert c, \bv_\cT) d\bz + \log q(c \vert \bw_\cC) \nonumber \\
 & \;\;\;\;\; \ge \sum_{i\in \cT}\bbE_{q(\bz\vert c, \bv_\cT)} \left[\log p(\by_i \vert \bx_i, c,\bu_\cC, \bz) \right] - \KL( q(\bz \vert c, \bv_\cT) \Vert q(\bz \vert c, \bv_\cC) ) + \log q(c \vert \bw_\cC) =: \cL_{\sf L}(\btheta; c, \cD), \label{eq:Ll}
\end{align}
with $\bv_\cT = \frac{1}{|\cT|}\sum_{i\in \cT} v(\cD_i)$. This bound will be used shortly to establish an integrated lower bound for both cases with or without label information. Note that we will abuse the notation $\btheta$ to denote the collection of all parameters in the encoders and decoders to be estimated.

\paragraph{Modeling a CM signal without label information.} Now, we discuss the case of units without label information. As we do not have labels, $c$ is regarded as a latent variable that needs to be estimated. The marginal likelihood becomes
\begin{equation}\label{eq:p_y_unlab}
    p(\by_\cT \vert \bx_\cT, \cD_\cC) := \int p(\by_\cT\vert \bx_\cT, c, \bu_\cC, \bz)q(\bz\vert c, \bv_\cC)q(c\vert \bw_\cC)d\bz dc.
\end{equation}
Note that Eq. \eqref{eq:p_y_unlab} is different from Eq. \eqref{eq:p_y_lab} in that it marginalizes out $c$. The lower bound of \eqref{eq:p_y_unlab} is derived by
\begin{align}
   \log p(\by_\cT \vert \bx_\cT, \cD_\cC) &:= \log \int \frac{p(\by_\cT\vert \bx_\cT, c, \bu_\cC, \bz)q(\bz\vert c, \bv_\cC)q(c\vert \bw_\cC)}{q(\bz\vert c, \bv_\cT)q(c\vert \bw_\cT)}q(\bz\vert c, \bv_\cT)q(c\vert \bw_\cT)d\bz dc \nonumber\\
& \; \ge -\sum_{l=1}^{L} q(c = l\vert \bw_\cT)\cL_{\sf L}(\btheta; c, \cD) + \bbH(q(c\vert \bw_\cT)) =: \cL_{\sf U}(\btheta; \cD),\label{eq:Lu}
\end{align}
where $\bbH(\cdot)$ denotes the entropy, that is, $\bbH(-q(c\vert \bw_\cT)) := -\sum_{l=1}^{L}q(c=l\vert \bw_\cT)\log q(c=l\vert \bw_\cT)$; $\bw_\cT$ is calculated similarly to $\bw_\cC$ but for $\cD_\cT$. 

\paragraph{The integrated model and its estimation.} Thus far, we have derived the lower bound \eqref{eq:Ll} and \eqref{eq:Lu} for a single arbitrary unit with and without a label, respectively. Now, let's build an integrated lower bound for the entire dataset $\cD^{\sf \cJ}$ for all historical units $j \in \cJ$ that encompasses both labeled and unlabeled units. This is expressed as 
\begin{equation}\label{eq:Llu}
    \cL_{\sf LU}(\btheta; \cD^{\sf \cJ}) = \sum_{j\in\cJ_{\sf L}}\cL_{\sf L}(\btheta; c^j, \cD^{j}) + \sum_{j\in\cJ_{\sf U}} \cL_{\sf U}(\btheta; \cD^j).
\end{equation}
The estimation of $\btheta$ is done by solving $\max_\btheta\cL_{\sf LU}(\btheta; \cD^{\sf \cJ})$ using stochastic optimization with the help of the reparameterization trick \cite{kingma2013auto}. In practice, a set of units $\cJ_{\sf L}^B \subset \cJ_{\sf L}$ and $\cJ_{\sf U}^B \subset \cJ_{\sf U}$ is randomly sampled to form a batch $\cJ^B := \cJ_{\sf L}^B \cup \cJ_{\sf U}^B$ at each optimization iterate to approximate \eqref{eq:Llu}. In doing so, one can randomly split $\cD^j$ into $\cD^{j}_\cT$ and $\cD^{j}_\cC$ for each unit $j\in \cJ^B$.





\paragraph{Joint prediction.} After training \model{} on the historical data, \model{} can make a joint prediction for the CM signal trajectory and the label for in-service unit $r$. Given available online observations $\cD_\cC^{r}$ as context, the label is inferred by $ \hat c^r = \argmax_l \{\bphi^r_l \}_{l=1,...,L}$ where $[\bphi^r_l]^\top_{l=1,...,L}$ is the output vector of $\phi(\bw^r_\cC)$ with $\bw^r_\cC = \frac{1}{\vert \cC^r \vert}\sum_{i\in\cC^r}w(\cD_i^r)$. The predicted mean and variance of $\by_*$ at the arbitrary input $\bx_*$ are calculated by 
\begin{equation}\label{eq:pred}
    \bbE[\by_*] \approx \hat \by_*\text{;}
\;\; \var[\by_*] \approx \frac{1}{K} \sum_{k=1}^{K} \hat \by_{*k}^2 - \left(\frac{1}{K} \sum_{k=1}^{K} \hat\by_{*k}\right)^2 + \hat\bsigma^{2}_*,
\end{equation}
with $\hat\by_* = \mu_{\sf d}(\bx_*, \hat c^r, \bu^r_\cC, \bmu_{\sf e}^r)$ and $\hat\bsigma^{2}_* = \sigma_{\sf d}(\bx_*, \hat c^r, \bu^r_\cC, \bmu_{\sf e}^r)$ with $\bmu_{\sf e}^r=\mu_{\sf e}(\hat c^r, \bv^r_\cC)$, where $\bu^r_\cC$ and $\bv^r_\cC$ are representations of $\cD^r_\cC$ encoded by $u(\cdot)$ and $v(\cdot)$, respectively. 
Here $\hat \by_{*k}$ can be calculated similarly to $\hat\by_{*}$ yet a sample $\bz^r_k \sim q(\bz\vert \hat c^r, \bv^r_\cC)$ is passed through the decoder instead of $\bmu_{\sf e}^r$, that is, $\hat\by_{*k} = \mu_{\sf d}(\bx_*, \hat c^r, \bu^r_\cC, \bz^r_k)$. One can interpret $\var[\by_*]$ as that $\frac{1}{K} \sum_{k=1}^{K} \hat \by_{*k}^{2} - \left(\frac{1}{K} \sum_{k=1}^{K} \hat\by_{*k}\right)^2$ captures the uncertainty imposed by encoding contexts to the latent space, while $\hat\bsigma_{*}^2$ estimates inherent noise in the observations \cite{kendall2017uncertainties}. Note that, in case label information $c^r$ is known, we can simply replace $\hat c^r$ with $c^r$ in the prediction process above.



\subsubsection{Modeling advantages} \model{} has critical advantageous properties as a means for predictive modeling of CM signals. 

\paragraph{Fast adaptation.} CM signals are often acquired in real-time. It necessitates instantaneous updates on the predictive model to incorporate newly arrived data on the fly. \model{} achieves this very efficiently without model retraining, by simply passing a new set of online observations $\cD^r_\cC$ through the encoders and decoder already learned. 
    
\paragraph{Label-awareness.} Our label-aware modeling seamlessly integrates label inference and regression within the VI framework. This facilitates enhancing signal predictions for units using their label information (e.g., system abnormality/status) or identifying their labels based on signal evolution when the labels are missing. 

\paragraph{Representation power and flexibility.} Modeling CM signals assuming a parametric form is becoming more challenging due to the increased complexity and non-stationarity of systems nowadays. By parameterizing distributions by DNNs, \model{} is endowed with strong representation power and flexibility that facilitate learning a wide range of function-generating distributions.

\paragraph{Uncertainty estimation.} In CM-based reliability engineering, quantifying uncertainty in predictions is crucial for subsequent decision-making, such as maintenance planning or inventory management. \model{} that includes a latent variable can capture inherent measurement noises and also uncertainty in the encoding process.

\subsection{Practical considerations}\label{sec:prac}

We discuss some considerations for practitioners to implement our proposed approach in practice.

\paragraph{Sampling contexts in training.} Typically, inputs from CM signals represent observation time points. Assuming that signals are functions defined over the input domain $[0, \tau]$, our setting involves estimating the function $f^r$ for in-service unit $r$ across the entire domain $[0, \tau]$ based on its online observations $\cD^r_\cC$ where $\bx_i^r \in [0, \tilde \tau]$ with the latest monitoring time $\tilde \tau < \tau$. Thus, employing a na\"ive random sampling approach to create contexts $\{\cD^j_\cC\}_{j\in\cJ}$ distributed throughout the entire domain $[0, \tau]$ during training does not reflect online observations of unit $r$ collected up to a specific time, often leading to inefficient learning.
A simple solution is drawing $\tau^* \sim \unif(0, \tau)$ at each optimization iterate and randomly choosing a certain number of observations available in $[0, \tau^*]$ to form contexts. As such, this context sampling strategy allows the model to learn how to predict the entire curve over $[0, \tau]$ based on online observations collected till a certain time point. 

\paragraph{A regularized objective function.} To further enhance learning of the inference network for labels $q(c\vert \bw_\cC)$, we can add a term to regularize the objective function \cite{kingma2014semi}:
$$
\tilde \cL_{\sf LU}(\btheta; \cD^{\sf \cJ}) = \cL_{\sf LU}(\btheta; \cD^{\sf \cJ}) + \lambda \sum_{j\in \cJ_{\sf L}} \left[-\log q(c^j \vert \bw^j_\cC)\right], 
$$
which assigns an extra weight to $-\log q(c\vert \bw_\cC)$. This encourages learning $q(c\vert \bw_\cC)$ based on the labeled signals $j\in \cJ_{\sf L}$. In general, we found $\lambda=0.1$ works well, as in \cite{kingma2014semi}.

\paragraph{Attention modules.} Incorporating attention modules \cite{vaswani2017attention} into NPs is a popular way to alleviate a well-known underfitting issue in NPs \cite{kim2019attentive}. We place self-attention layers to the encoder DNNs $u(\cdot), v(\cdot),$ and $w(\cdot)$ after their initial fully-collected layers. Also, we further replace the averaging operator for $\bu_\cC$ with a cross-attention module across the context representations $\{u(\cD^j_i)\}_{i\in \cC^j}$ and a target input $\bx^j_i$ with $i\in \cT^j$; thus, $\bu_\cC$ depends on $\bx^j_i$. The self-attention module can produce a richer representation by modeling interactions within context points. Furthermore, the cross-attention module helps place stronger attention on the context points close to the location of the target point, encouraging the prediction at the target point $\bx^j_i$ to be closer to the outputs of the contexts nearby. Such a structure is adopted in our numerical studies in Sec. \ref{sec:simulation}. We provide a pictorial description of the attention module-integrated structure in Appendix A.

\paragraph{Functional data augmentation.} A key challenge in training NPs is the need for enough functional observations. The challenge comes from the nature of NPs, which directly learns a distribution over functions. In the context of CM, this implies that we may need many historical units to train \model{}, which may not always align with real-world scenarios. Such an issue becomes further challenging when CM signals exhibit highly nonparametric trends or heterogeneity across units, which requires more function samples to learn the underlying function-generating distribution. Appendix B and C provide a practical data augmentation framework based on functional data analysis to alleviate this issue, along with a simulation study demonstrating its usefulness.

\section{Related work}\label{sec:related work} 

A large body of literature exists on the predictive modeling of CM signals and NPs. As such, this section discusses studies particularly related to our work. For comprehensive reviews, please see the excellent review paper for predictive modeling in CM \cite{si2011remaining} and for the NP family \cite{jha2022neural}.

\paragraph{Personalized predictions for CM signals.}
One natural way to achieve personalized predictions for an individual CM signal is based on a two-step approach. The first step characterizes the general trend from historical CM data. Then, the trend is gradually refined to model an in-service unit individually. A popular approach uses a mixed-effect model \citep{gebraeel2005residual}. They estimate mixed-effect coefficients using historical CM signals. Then, the random effects are updated from online observations using empirical Bayes. Along this line, many extensions have been explored. Those include studies that account for heterogeneity in historical CM signals \citep[][]{yuan2014hierarchical}, external sources of system failures \citep[][]{gao2019reliability}, time-varying system degradation rates \citep[][]{si2014estimating}, and multiple phases in degradation processes \citep[][]{wen2018degradation, wang2020mixed}, to name a few. Note that the methods above assume a parametric form for signals or a linear mean degradation path \citep{kontar2018nonparametric}. While facilitating fast adaptation to online data, such parametric modeling results in severe vulnerability to model misspecification. To tackle this, another line of two-step approaches inspired by functional data analysis has been proposed \citep[][]{zhou2011degradation, chung2020functional, fallahdizcheh2022transfer}. They estimate eigenfunctions from historical CM signals and express each signal as a linear combination of the eigenfunctions. Then, personalization is done by updating the coefficients based on the online data. In contrast to parametric model-based approaches, the alternative methods do not impose a restriction on CM signals to be modeled as a parametric form or a monotonic mean trend. Unfortunately, these approaches suffer from limited scalability in the number of historical units.

Alternatives to two-step modeling directly derive a predictive distribution given both historical and online observations. In this regard, our model falls within this category. A notable approach in this category is based on multi-output Gaussian processes (GPs) \citep{kontar2018nonparametric}. It pools all training and testing CM signals into a large multivariate GP, through which information transfer occurs from the historical to the testing units. Based on this idea, multiple studies further improve the model by incorporating auxiliary data \citep{kontar2017nonparametric}, adaptively selecting more informative units \citep{wang2022regularized}, and distributing model inference efforts to individual units while preserving their privacy \citep{chung2023federated}, among others. They all are capable of personalized predictions using non-parametric modeling, a key benefit inherited from GPs. However, their personalization process introduces a substantial latency for online updates, as it requires model retraining each time a new data point is obtained. Needless to say, this is a tedious task, especially for high complex models such as GPs. Recent studies propose online learning schemes for multi-output GPs to avoid model retraining from scratch \citep[e.g.,][]{yang2018online, hu2021nonlinear}. Nonetheless, their iterative and nonparallelizable nature greatly impedes their use when the number of online observations is excessive.

\paragraph{NPs and their applications.} Recently NPs have drawn significant attention due to their rich expressiveness, scalability in both training and test stages, and the ability to learn meta-representations \citep{jha2022neural}. After their first introduction by \cite{garnelo2018conditional}, multiple studies have made extensions in various directions. For example, studies aim to capture global uncertainties in the encoding process \citep{garnelo2018neural, lee2020bootstrapping}, mitigate underfitting to the context points \citep{kim2019attentive, kim2022neural, nguyen2022transformer}, account for correlations across functions \citep{singh2019sequential, yoon2020robustifying} or across target output points \citep{markou2021efficient}, ensure invariant predictions under input shifts or transformations \citep{gordon2019convolutional}. Built upon the methodological advances above, NPs have been widely used in a range of domains, including clinical data analysis \citep{kia2019neural}, climate science \citep{vaughan2021convolutional}, image classification \citep{wang2022np}, robotics \citep{chen2022meta, li2022category}, and so on. Despite that, their applications in the predictive modeling of CM signals have yet to be investigated.




In the sense of using incomplete label information, there are few recent studies relevant to \model{}. A study proposed an NP-based model, called SNPAD, for anomaly detection using partially labeled data \cite{zhou2023semi}. SNPAD includes an encoder that maps both labeled and unlabeled observations into a latent space, followed by a decoder that estimates an anomaly score of each observation. Another related study in this regard proposes NP-Match  \cite{wang2022np}, an NP-based model designed for semi-supervised image classification tasks where labeled and unlabeled observations are regarded as contexts and targets, respectively. While NP-Match, SNPAD, and \model{} consider both labeled and unlabeled observations simultaneously, a direct comparison is not straightforward. This is due to their different objectives and settings where \model{} (i) focuses on real-time personalization in CM and (ii) provides joint predictions for functions and their labels, whereas SNPAD and NP-Match aim to predict anomaly scores or labels only.

\section{Model validation}\label{sec:model_val}

In this section, we design and discuss numerical studies to validate our proposed model using both simulated and real-world data. Sec. \ref{sec:simulation} assesses our proposed approach using simulated data. Sec. \ref{sec:case_study} tests our model on a reliability engineering application about CM-based prognosis for lithium-ion batteries. We note that Appendix C includes an additional simulation study that validates our proposed functional data augmentation scheme. 

\subsection{Simulation study}
\label{sec:simulation}

Our simulation study is designed to answer the following questions. (i) Can \model{} leverage label information to predict heterogeneous future trends of signals when early-stage online data is insufficient to reveal the future heterogeneity?  (ii) Can \model{} outperform benchmark models over scenarios where varying proportions of observations are labeled?

\paragraph{Benchmark models.} We compare \model{} with the benchmark models denoted as follows. 
\begin{itemize}
    \item \anp{}: an attentive NP model \cite{kim2019attentive}. This model builds upon $\lnp{}$, including both self- and cross-attention modules. The comparison of \model{} with \anp{} will highlight the benefit of leveraging labels. 
    \item \mgp{}: a variational multi-output GP model \cite{alvarez2008sparse}. This model builds cross-correlation between units based on convolution processes. Note that we chose a variational version of multi-output GPs to make it scalable to our numerical study. When online data arrives, \mgp{} needs retraining for an updated prediction. 
\end{itemize}
For a fair comparison, \anp{} has a similar network configuration to \model{}, where both are equipped with attention modules and $d_{\sf lat} = 8$. Detailed settings can be found in Appendix A. 



\begin{figure}[h!]
 \centering
     \includegraphics[width=\columnwidth]{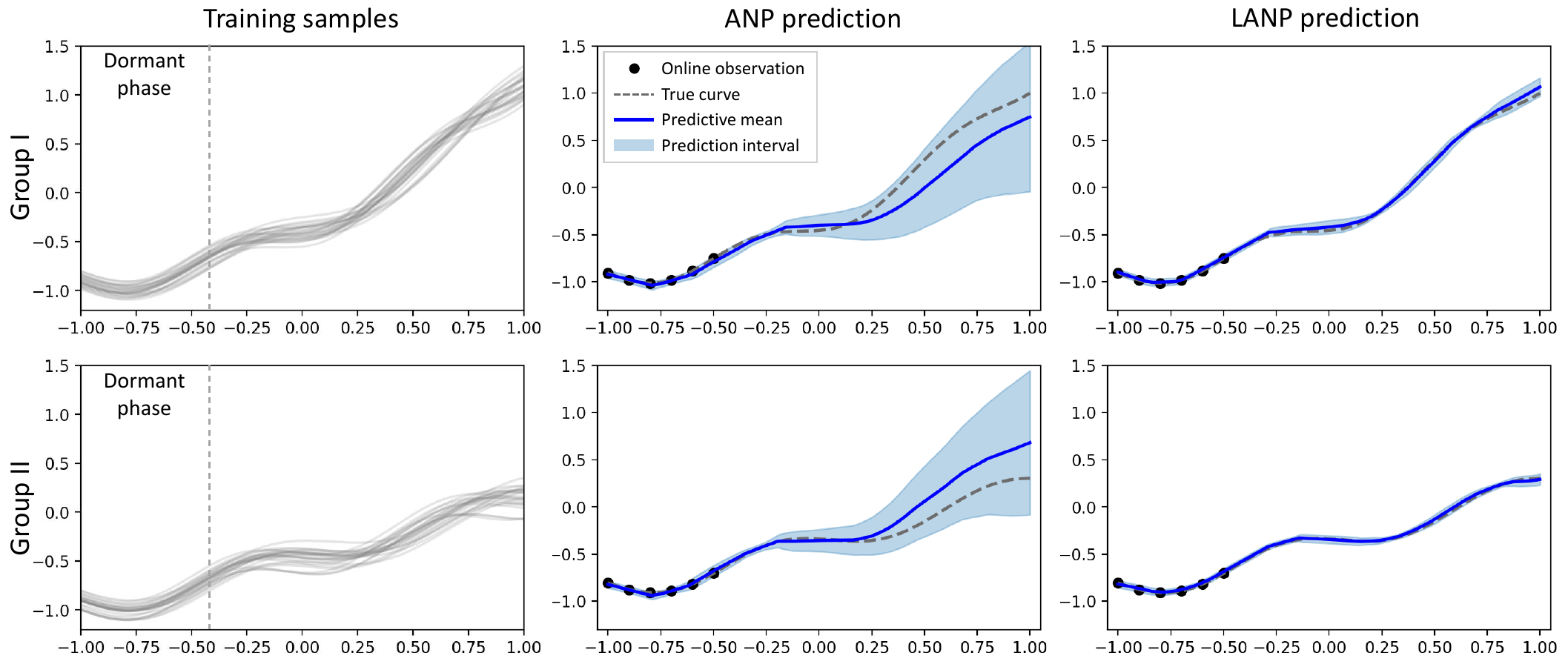}
 \caption{Predictions by \anp{} and \model{} $(\alpha=0.3, \gamma=0.25)$.   
}\label{fig:simul1}
 \end{figure}

\paragraph{Setup.} We consider two groups of signals generated from data generation models:
\begin{itemize}
    \item \text{Group I:} $ y = 0.3 x^2 -2 \sin(b_1 \pi x) + b_2 + \epsilon, \; x \in (3,10].  $
    \item \text{Group II:} $y = \begin{cases} 0.3 x^2 -2 \sin(b_1 \pi x) + b_2 + \epsilon, \;\;\; x \in (0,3], \\  1.8 x^2 -2 \sin(b_1 \pi x) + b_2 -2.7 + \epsilon, \;\;\; x \in (3,10], \end{cases}$
\end{itemize}
with $b_1 \sim \unif(0.35, 0.45)$, $b_2 \sim \unif(0, 3)$, and the white noise $\epsilon \sim \cN(0,0.03^2)$. Samples of the signals are displayed in the first column in Fig. \ref{fig:simul1}. It is noteworthy that both data-generating models share a common underlying function in the range $x\in(0,3]$. We call this as the \textit{dormant stage}. This is common in CM practices, where the future evolution of CM signals is dependent on different operational modes or configurations (i.e., labels), but their heterogeneity is not easily detected in the early stage \cite{chung2020functional}. Predicting a future trend solely based on online observations at $x\in (0,3]$ would be very challenging in such a case. By comparing \model{} with benchmark models, we can evaluate if \model{} can effectively leverage label information and, therefore, accurately predict future heterogeneous trends. 

We train \model{} and \anp{} throughout 25,000 iterations. At each iteration we generate a batch of 16 signals, comprised of 8 signals from each group, to train the models. We consider partially labeled cases by randomly removing group labels for $16(1-\gamma)$ signals for $\gamma \in \{0, 0.25, 0.5, 0.75, 1\}$. That is, setting $\gamma = 0$ leads to the case without any group information, whereas $\gamma = 1$ represents that the group information of all signals is available. A generated signal comprises 45 observations. To form contexts for the signal, a random number of observations greater than 2 and less than 15 are selected in a way described in Sec. \ref{sec:prac},  while the remainder is set to target observations. Meanwhile, we generate 100 training signals from each group to train \mgp{}. This restriction is imposed due to the rapid unscalability of \mgp{} as $J$ increases. Yet, we observed that training \mgp{} on more than 100 training signals does not lead to a substantial increase in its predictive performance, and thus a fair comparison is still available. We validate the models on 20 test signals for each group. For a test signal, we generate 20 observations evenly spaced over $x \in (0,10)$. We consider the ``$\alpha$-degradation" stage: where observations in the first $100\alpha \%$ of the entire trajectory are collected as online data that forms contexts in NP-based models. Evaluation for model predictions is based on the root mean squared of errors (RMSE) metric at 400 points evenly spaced over $x \in (0,10)$.

\paragraph{Results.} Now we discuss experimental results. Table \ref{tab:simul1-rmse} includes average RMSEs over 20 test signals for the compared models with $\alpha = 0.3, 0.5, 0.7$. Based on the results, we can obtain important insights. 
Specifically, \model{} significantly outperforms \anp{} when $\gamma > 0$, that is, label information is provided. This demonstrates \model{}'s capability of label-awareness that contributes to more accurate signal prediction. Fig. \ref{fig:simul1} illustrates the predictions of \model{} and \anp{} for two representative signals from Group I and II, when current online observations are not sufficient to distinguish signals from different groups. As shown, \model{} can distinguish heterogeneity in the future evolution based on label information, while \anp{} fails to do so. We also emphasize that such merits in \model{} can be achieved even with a small portion of labeled signals in the training dataset ($\gamma=0.25$). 
Meanwhile, it is not surprising that the predictive accuracy of \model{} improves as more online observations or labeled signals are provided. 

\begin{table}[h!] 
\centering
\caption{Average RMSEs over different degradation stages (group I).}
\begin{tabular}{c|c|ccc}
\hline
Models & $\gamma$    & $\alpha=0.3$        & $\alpha=0.5$        & $\alpha=0.7$              \\ \hline
\mgp{}  & -& 0.406 (0.181) & 0.226 (0.200) & 0.189 (0.045) \\\hline
\anp{}    & -& 0.187 (0.019) & 0.106 (0.090) & 0.025 (0.012) \\\hline
\multirow{5}{*}{\model{}}  & 0& 0.245 (0.027) & 0.144 (0.094) & 0.076 (0.020) \\
& 0.25& 0.022 (0.011) & 0.017 (0.003) & 0.011 (0.004) \\
  & 0.5& \textbf{0.014} (0.005) & 0.009 (0.002) & 0.011 (0.002) \\
  & 0.75& 0.018 (0.007) & 0.009 (0.002) & 0.009 (0.002) \\
  & 1& 0.020 (0.011) & \textbf{0.007} (0.002) & \textbf{0.007} (0.003) \\\hline
\end{tabular}  \label{tab:simul1-rmse}
\end{table}

\begin{figure}[t!]
 \centering
     \includegraphics[width=0.7\columnwidth]{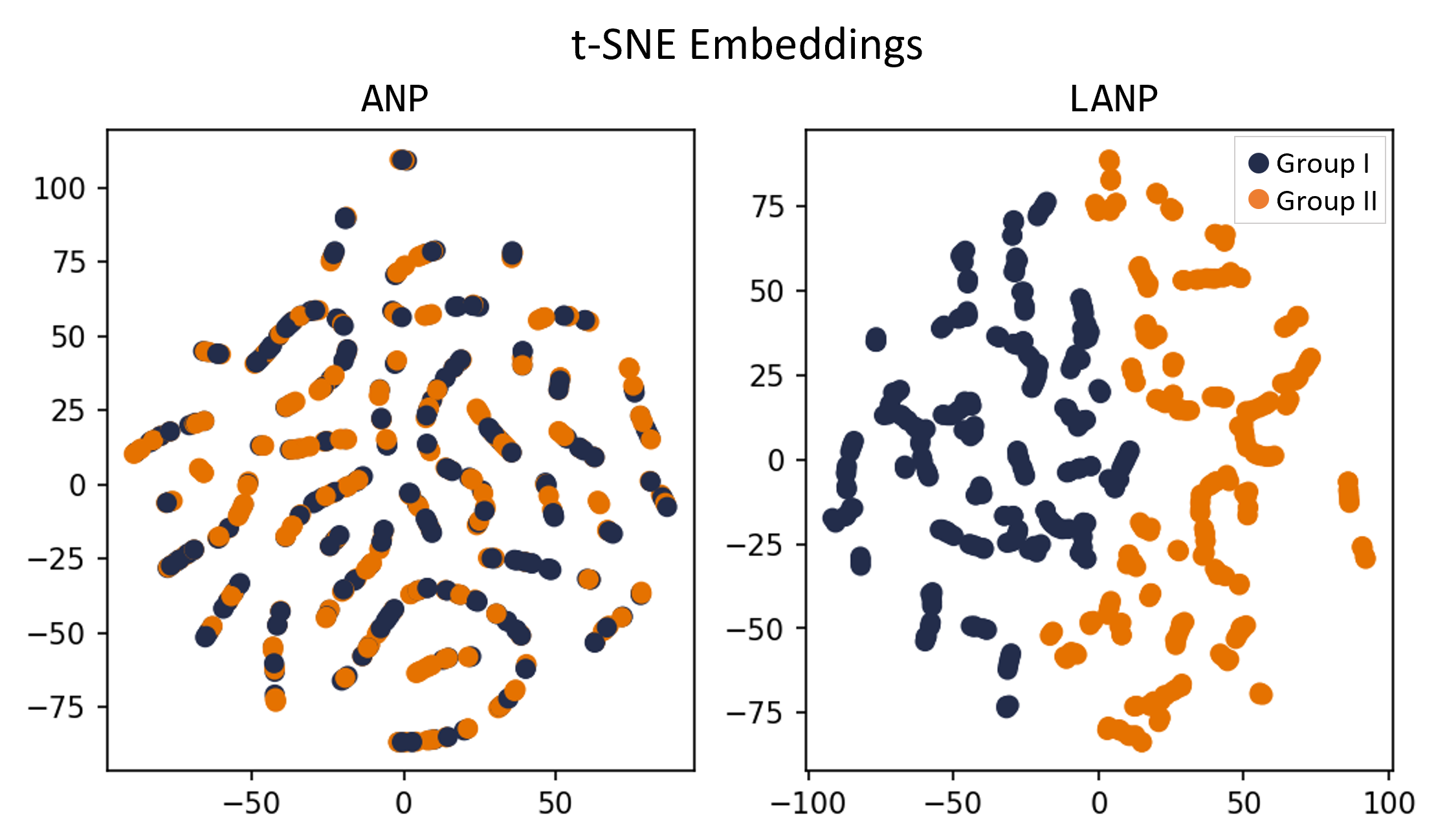}
 \caption{t-SNE embeddings of representations at the dormant stage $(\alpha=0.3)$.  
 \vspace{-1pc}
 }\label{fig:tsne}
 \end{figure}

To investigate the label-awareness ability of \model{}, Fig. \ref{fig:tsne} depicts 2-dimensional embeddings of the means of $\bz$ in the representation space. We generate online observations of 100 test signals from each group at the dormant stage and plot the t-SNE embeddings \cite{van2008visualizing} of their 8-dimensional representations encoded by \anp{} and \model{}. We directly see from Fig. \ref{fig:tsne} the means of $\bz$ encoded by \anp{}  (i.e., $\mu_{\sf e}( \bv^r_\cC)$) are not clustered by different groups, as signals are not distinguishable with their online observations at the dormant stage. On the other hand, \model{} can cluster the means of $\bz$ (i.e., $\mu_{\sf e}(c^r, \bv^r_\cC)$) by their groups using their label information $c$. Such clustering in the representation space results in distinguished decodings by groups. Consequently, \model{} is able to differentiate predictions for future trajectories according to their groups.

Table \ref{tab:simul1-time} provides times in seconds taken by each model to update its prediction using online observations. We can observe that \model{} and \anp{} dramatically reduce the time for update compared to \mgp{}. This sheds light on the critical ability of \model{} to facilitate real-time inference in online regimes. \model{} does not require retraining at runtime when adapting to new online observations. Instead, the adaptation process simply passes new observations through already trained DNNs and thus is parallelizable. As such, real-time adaptation remains scalable even when many online observations are introduced simultaneously. This highlights \model{}'s potential within a range of real-world CM applications where real-time updates are instrumental. In addition to these merits, \model{} comes with strong representation power resulting from the parameterization using DNNs, which in turn results in more accurate signal predictions as shown in Table \ref{tab:simul1-rmse}. Finally, we note that an additional simulation study on the functional data augmentation scheme is provided in Appendix C.

\begin{table}[htbp!]\centering
\caption{Times in seconds for online updates. Figures indicate times averaged through the updates for 20 online observations.}
\begin{tabular}{c|ccccc}
\hline
     & \mgp{}      & \anp{}         & \model{}       \\ \hline
Time & 58.43 (0.05) & 0.0033 (0.00) & 0.0051 (0.00)       \\ \hline
\end{tabular}\label{tab:simul1-time}
\end{table}

\subsection{Case study: data-driven prognosis and anomaly detection for lithium-ion batteries}

The case study presents an application of \model{} in the prognosis of lithium-ion batteries, where a decrease in capacity characterizes battery degradation. The reliable operation of systems/devices with a lithium-ion battery hinges on timely maintenance or replacement. In this regard, it is crucial to achieve an accurate estimation of future degradation trends and early detection of anomalies. To this end, data-driven approaches build a prediction model based on data from both historical batteries and the in-service battery. A critical challenge is that capacity degradation of the in-service battery is often collected in real-time through advanced IoT sensors. Hence, the capability of rapid adaption to incoming data streams is imperative.

\paragraph{Dataset.} We use the CALCE battery anomaly detection dataset \cite{lee2018reduction}. It contains capacity decrease trajectories obtained from 23 battery cells. Among them, 14 batteries are regarded as representative of the qualified production lot, whereas 9 batteries are categorized as disqualified. That is, the label of a battery denotes its qualification. While initially presenting as healthy, the disqualification is inherently manifested at the later stage as abnormal degradation trends unacceptably deviating from a normal trend. Details can be found in \cite{lee2018reduction}. 

\paragraph{Setup.} In our experiment, we carry out leave-one-out cross-validation to evaluate models. Thus, we assess models over 23 cases, each choosing one battery cell for testing. We aim (i) to forecast the future evolution of the in-service battery's capacity and (ii) to determine its qualification status (qualified or not). The benchmark models take a two-stage approach to detect anomalies: estimating a degradation curve, and categorizing it whether qualified or not using a classification algorithm. We adopt k-nearest neighbors (KNN) or support vector machine (SVM) as the classification algorithm. We also augment the training data using the proposed data augmentation method, discussed in Appendix B. 

\begin{figure}[h!]
 \centering
     \includegraphics[width=\columnwidth]{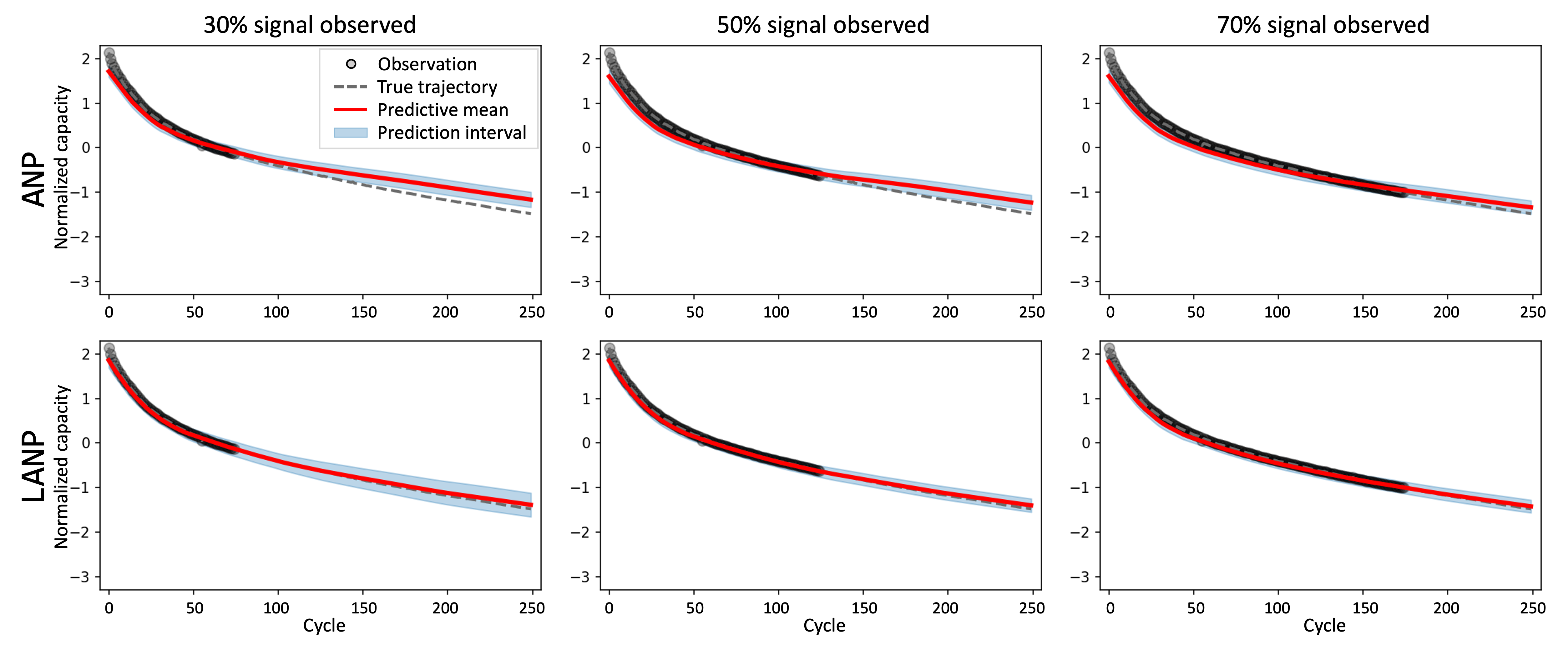}
 \caption{CM signal predictions over different proportions of available online observations (battery cell ID: \#23).}\label{fig:calce-pred}
 \end{figure}

\begin{figure}[h!]
 \centering
     \includegraphics[width=0.5\columnwidth]{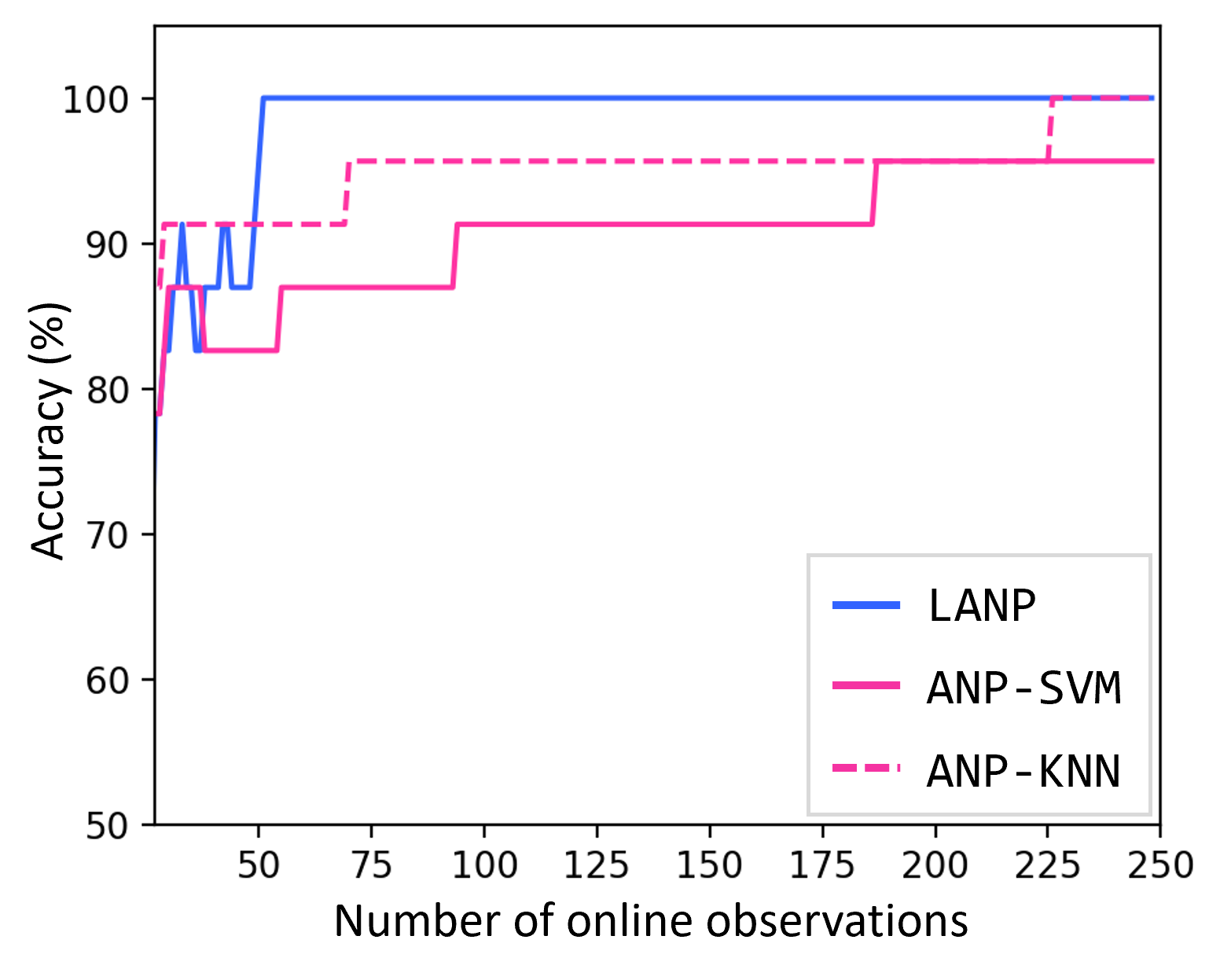}
 \caption{Accuracy of label prediction over different numbers of online observations. \texttt{ANP-SVM} denotes the benchmark model that first estimates signals using \texttt{ANP-KNN} and then classifies them using SVM to determine qualifications. Similar notation applies to \texttt{ANP-KNN}. }\label{fig:calce-label} \vspace{-1pc}
 \end{figure}

\paragraph{Results.} Results are presented in Table \ref{tab:calce-rmse}, Fig. \ref{fig:calce-pred}, and Fig. \ref{fig:calce-label}. We see that, \model{} outperforms \anp{} in curve predictions over different degradation phases. It demonstrates that our label-awareness modeling leads to better learning of the function-generating distribution. This comes with well-quantified predictive uncertainties, as presented in Fig. \ref{fig:calce-pred}. Furthermore, Fig. \ref{fig:calce-label} presents that \model{} improves detecting anomalies of batteries over the benchmark models. This reveals the advantage of our joint estimation for curves and labels, compared to the two-stage approaches. It is particularly promising to see \model{}'s ability to detect anomalies at an early degradation stage. Importantly, such advantages of \model{} come along with its real-time adaptability. This further highlights the strong potential of \model{} in reliability engineering for advanced systems with real-time CM data.


\begin{table}[h!]
\centering
\caption{Average RMSEs of future signal predictions given different proportions of available online observations.}
\begin{tabular}{c|ccc}
\hline
Models & $\alpha=0.3$            & $\alpha=0.5$            & $\alpha=0.7$            \\ \hline
\mgp{}                    & 0.0912 (0.0814)           & 0.0693 (0.0391)           & 0.0501 (0.0321)           \\
\anp{}                     & 0.0922 (0.0936) & 0.0616 (0.0495) & 0.0493 (0.0354) \\ 
\model{}                    & \textbf{0.0824} (0.0611) & \textbf{0.0560} (0.0377) & \textbf{0.0469} (0.0318) \\ \hline
\end{tabular}\label{tab:calce-rmse}
\end{table}

\label{sec:case_study}

\section{Conclusion}
\label{sec:conclusion}

The successful deployment of modern digital-twin technologies and cyber-physical systems hinges upon real-time adaptation to system changes. Endeavors to develop predictive models capable of rapid prediction updates on real-time online data are often challenged by the trade-off between representation power and agility to new data. Our study is able to tackle the trade-off, achieving both capabilities simultaneously.  

This paper presents an approach for instantaneous updates of predictions personalized to individual units based on their online CM data. Building upon the NP framework, our model can attain both real-time prediction updates and strong representation power. This is achieved along with improved predictions through label-aware modeling that enables the use of label information of curves that may be incomplete, as well as inference of labels in case of missingness.
Numerical studies demonstrate that our model is capable of accurate curve predictions with well-quantified uncertainty, joint inference of curves and labels, and real-time updates of predictions based on online observations.

\Urlmuskip=0mu plus 1mu
\bibliographystyle{unsrt}
\bibliography{references}

\appendix

\setcounter{equation}{0}
\setcounter{figure}{0}
\renewcommand{\thefigure}{A.\arabic{figure}}
\renewcommand{\theequation}{A.\arabic{equation}}

\section*{APPENDIX}
\section{Model configuration}

\begin{figure}[h!]
 \centering
     \includegraphics[width=.65\textwidth]{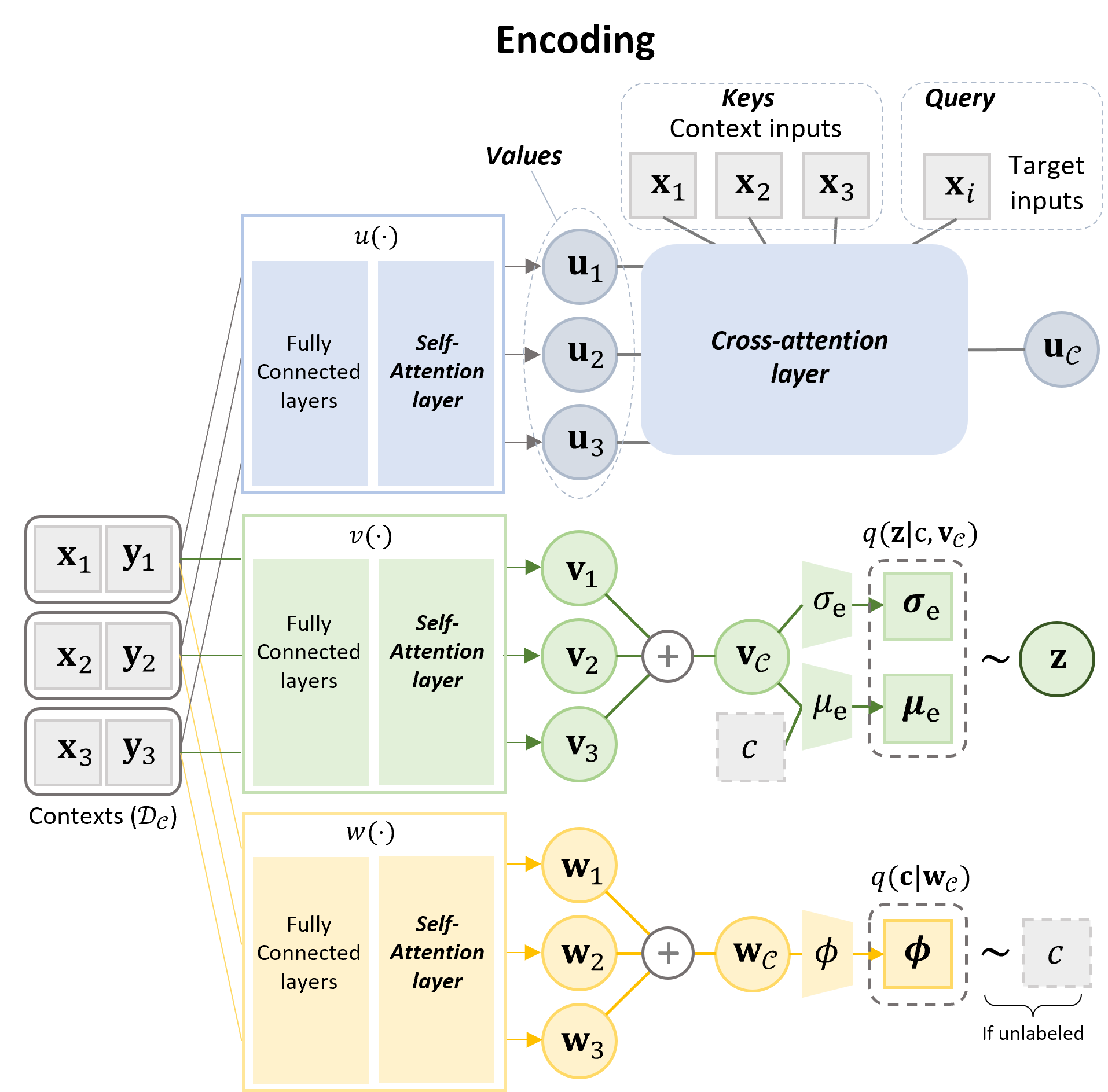}
 \caption{The encoders of \model{} with attention modules used in numerical studies.}\label{fig:attn}
 \end{figure}

Fig. \ref{fig:attn} illustrates the configuration of \model{}'s encoders equipped with attention modules \citep{vaswani2017attention} used in the numerical studies. Note that all notations but functions ($u, v, w, \mu_{\sf e}, \sigma_{\sf e}, \phi$) in the figure omit the index $j$. The encoders $u(\cdot), v(\cdot),$ and $w(\cdot)$ start with two fully-connected layers, followed by uniform self-attention modules. While $\bv_\cC$ and $\bw_\cC$ are the aggregation of representations by the averaging operator, $\bu_\cC$ is obtained by a multi-head cross-attention module across contexts and the target input. Specifically, the representations $\{\bu_i\}_{i\in\cC}$, context inputs $\{\bx_i\}_{i\in\cC}$, and a target input $\bx_i$ serve as values, keys and a query in the cross-attention module. Other encoding DNNs $\sigma_{\sf e}(\cdot)$, $\mu_{\sf e}(\cdot)$, $\phi(\cdot)$ consist of a single fully-connected layer. We set $d_{\sf lat} = 8$ for the dimension of $\bu_\cC$ and $\bz$. The decoding DNNs $\sigma_{\sf d}(\cdot)$ and $\mu_{\sf d}(\cdot)$ consist of two fully-connected layers. All fully-connected layers have 128 nodes.

\section{Functional data augmentation}\label{sec:fda}

This section discusses our proposed data augmentation framework for historical CM signals based on a statistical method for functional data analysis, called functional principal component analysis (FPCA) \citep{ramsay2nd}.

FPCA has a rich history in longitudinal data analysis. It is an extension of PCA to a functional data setting, where each observation is a function (or a curve) rather than a single vector. Conceptually, it can be thought of as dimensionality reduction for functional data, which encodes functions to the corresponding finite-dimensional vectors in the space ``spanned'' by a finite number of orthonormal basis functions \citep{chung2020functional}. Here the projected vectors and the basis functions are called functional principal component (FPC) scores and eigenfunctions, respectively. To apply FPCA to our historical CM signals, let $y^j(x)$ denote the observation of unit $j$'s CM signal at time $x \in \cP \subset \bbR$ where $\cP$ indicates a time domain, respectively. FPCA approximates $y^j(x)$ as
\begin{equation}\label{eq:fpca}
    y^j(x) \approx \hat \beta(x) + \sum_{n=1}^N \hat \xi_{jn} \hat \psi_n(x) + \eps(x), \;\;\;\;\;\;\;\; \text{for } j\in \cJ,
\end{equation}
where $\hat \beta(x)$ represents an estimated mean trend across $y^1(x),...,y^J(x)$, $\hat \bxi^j := [\xi_{j1},...,\xi_{jN}]^\top \in \bbR^N$ is an estimated FPC score vector for unit $j$, $\hat \psi_n(x)$ represents the estimated $n$-th eigenfunction, and $\eps(x)$ is the \textit{i.i.d.} Gaussian noise. We note that there are multiple off-the-shelf packages for FPCA, such as \texttt{refund} \citep{refund} in R and \texttt{scikit-fda} 
\citep{ramos2022scikit} in Python. For the theoretical properties and detailed estimation processes of FPCA, the reader is referred to \cite{ramsay2nd}. 

\begin{figure}[h!]
 \centering
     \includegraphics[width=1\textwidth]{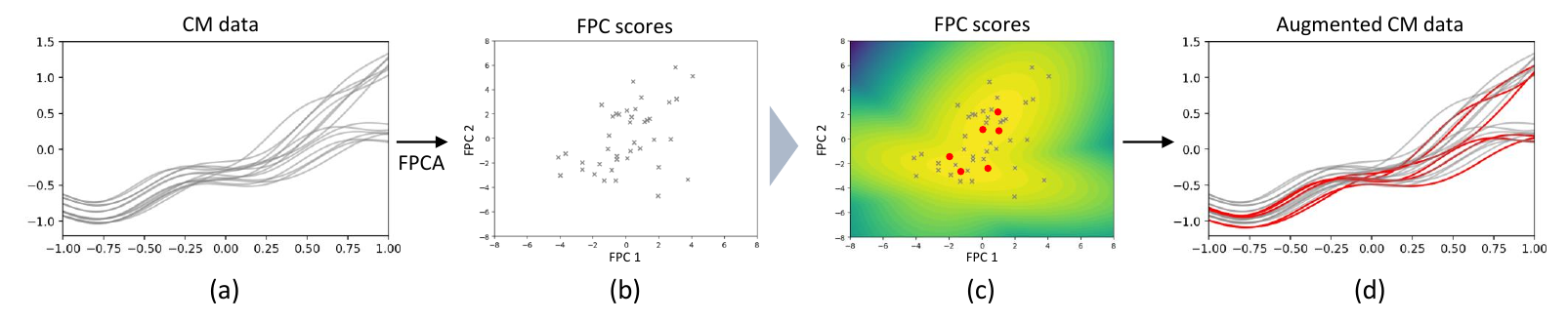}
 \caption{Signal augmentation by the proposed FPCA-based approach. (a) Original CM data. (b) FPCs for CM signals estimated by FPCA (cross marks). (c) A generative model fitted to FPCs and generated samples (red dots). (d) Augmented CM data with reconstructed synthetic signals (red solid lines).
 }\label{fig:fpca}
 \end{figure}

Our functional data augmentation approach is inspired by the crucial property of FPCA: given $\hat \psi_n(x)$ and $\hat \beta(x)$, the CM signal of unit $j$ is characterized by its FPC score estimate $\hat \bxi^j$. The key idea is to build a generative model that estimates the distribution of FPC scores $\{\hat \bxi^j\}_{j\in\cJ}$. Then we can draw an FPC score sample $\tilde \bxi = [\tilde \xi_n]^\top_{n=1,...,N} \in \bbR^N$ from the estimated generative model and reconstruct a function $\tilde y(x)$ corresponding to the sample using Eq. \eqref{eq:fpca}, written by $\tilde y(x) = \hat\beta(x) + \sum_{n=1}^N \tilde \xi_{n} \hat \psi_n(x)$. As such, our generation process does not directly adjust functions; rather, it generates samples in the FPC space and reconstructs their corresponding functions. Fig. \ref{fig:fpca} depicts the proposed functional data augmentation method. While any generative model can be employed, we use a Gaussian mixture model with $Q$ components. Specifically, we model the joint probability distribution of $\{\hat \bxi^j\}_{j\in\cJ}$ by
$$
p(\hat \bxi^1,...,\hat \bxi^J) = \prod_{j\in \cJ} p(\hat \bxi^j) = \prod_{j\in \cJ}\sum_{q=1}^{Q}\omega_q\cN(\hat \bxi^j ; \mathbf{m}_q, \bSigma_q)
$$
where the $q$-th component of the mixture model is characterized by a Gaussian distribution with the mean $\mathbf{m}_q$ and covariance $\bSigma_q$, weighted by $\omega_q \ge 0$ with $\sum_{q=1}^Q\omega_q = 1$. The parameters can be estimated by the expectation-maximization algorithm \citep{dempster1977maximum}. For more details on the estimation and sampling process for Gaussian mixture models, please refer to \cite{bishop2006pattern}. 

The proposed approach for functional data augmentation enjoys several notable advantages. Specifically, the estimated eigenfunctions and the distribution of FPC scores can capture an underlying structure and variability inherent across the historical signals. Thus, leveraging these eigenfunctions and the distribution of FPC scores enables the generation of realistic synthetic signals. Moreover, the generation process can be done in an efficient way. This is because the generative model builds upon FPC scores, which are low-dimensional representations of signals, rather than directly dealing with the signals. Finally, the approach generates smooth signals. This is a direct consequence of using FPCA, which estimates smooth eigenfunctions. Here we should note that the proposed approach is amenable to cases for one-dimensional input, e.g., temporal data. That being said, this method is highly practical in common CM applications, given that CM observations often manifest as time-series signals.

\paragraph{Model training with synthetic signals.} While exhibiting a realistic trend, synthetic signals do not have a label. We thus use them as though unlabeled signals. When training \model{} in practice, a batch at each iteration comprises labeled and unlabeled real signals, as well as unlabeled synthetic signals if needed.

\section{Additional simulation study}\label{sec:simul2}

Here, we validate our proposed data augmentation method based on FPCA. We assess the performance of \model{} trained on augmented data versus original data. We vary the parameter distribution of the data generation model to identify which case the proposed data augmentation approach excels. We evaluate models based on their predictions for signals and labels.  

\paragraph{Setup.} We consider two data generation functions as follows.
\begin{itemize}
    \item \text{Group I:} $ y = b_1 \cos(x) + 1.5x + b_2 + \epsilon, \hspace{3pc} x \in (0,10)$
    \item \text{Group II:} $ y = b_1 \sin(x) + 1.5x + b_2 + \epsilon, \hspace{3pc} x \in (0,10)$
\end{itemize}
where $b_1 \sim \unif (0.5, 1+\delta)$, $b_2 \sim \unif (0, 2+2\delta)$, and $\epsilon \sim \cN(0,0.03^2)$ denotes random white noise. We consider two cases by setting $\delta$ to two different values: $0.5$ and $2$. It is important to note that $\delta$ controls the difficulty of learning a function-generating distribution. For instance, a greater $\delta$ induces the generation of training samples that exhibit greater deviations. Inference of the function-generating distribution based on such samples is more challenging compared to cases with less deviating samples (i.e. with smaller $\delta$). In each of the two cases, we generate a dataset consisting of 16 partially-labeled training signals. This dataset comprises 8 training signals from each group, where we intentionally remove group labels for 4 signals selected at random. In contrast to Setting I, we have a limited training dataset of only 16 signals in this setting. This reflects possible real-world scenarios in CM applications where only a limited number of historical units are available, where data augmentation can help. During each iteration in training \model{}, we randomly select 8 labeled original signals and generate 8 synthetic signals by the method detailed in Sec. \ref{sec:fda}. We compare this approach with training \model{} solely on the original signals. Other configurations for data generation, models, and optimization remain the same in Setting I. We iterate the experiment 5 times.

\begin{figure}[h!]
 \centering
     \includegraphics[width=0.5\textwidth]{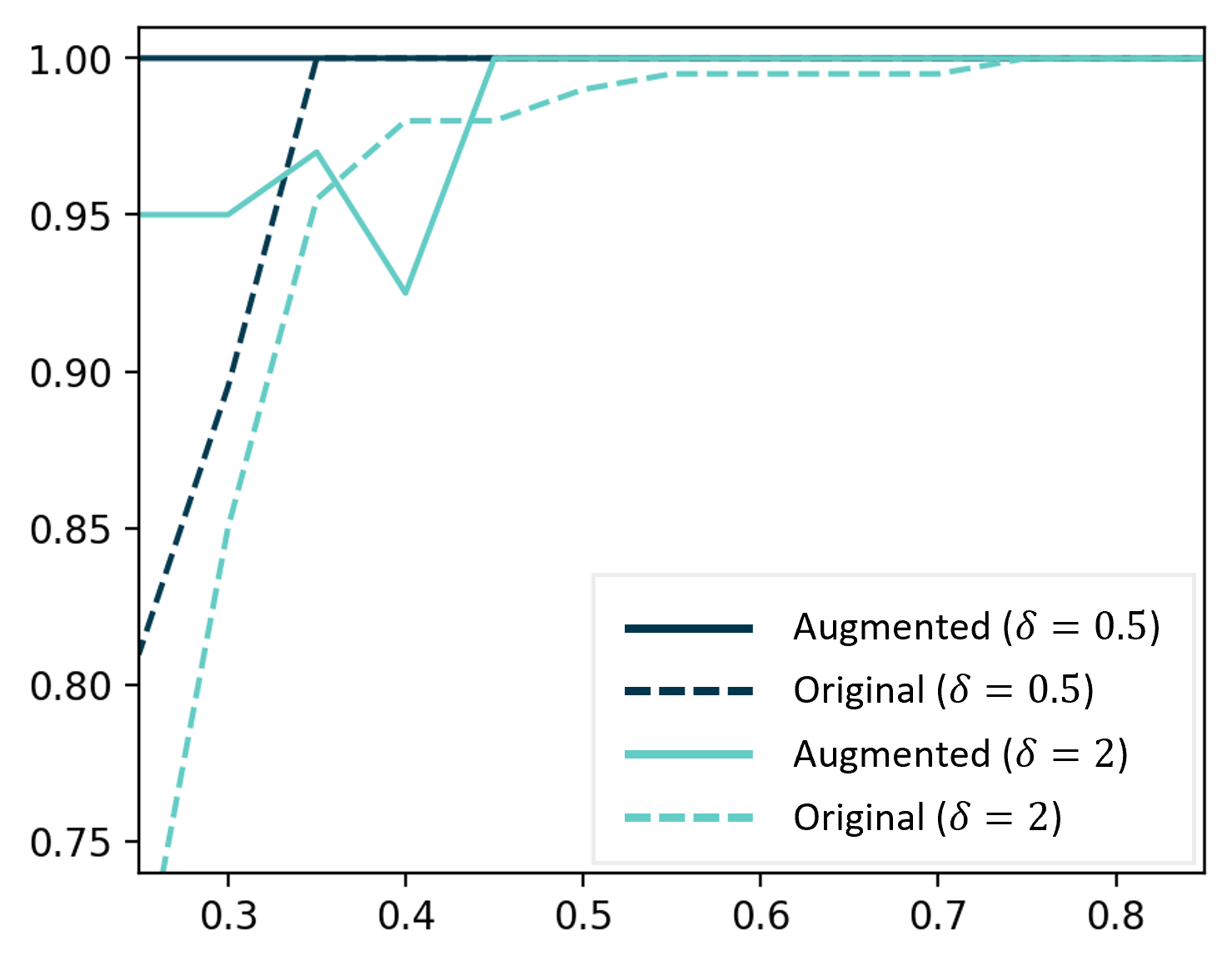}
 \caption{Accuracy of label prediction over different numbers of online observations.}\label{fig:setting2}
 \end{figure}

\paragraph{Results.} Table \ref{tab:fpca} shows results on average RMSEs for cases with $\delta=0.5$ and $\delta=2$. 
Fig. \ref{fig:setting2} shows label prediction accuracies over different numbers of online observations. 

\begin{table}[h!]\centering\caption{Average RMSEs of \model{} with or without data augmentation. In the first column, ``Original" and ``Augmented" refer to the cases where \model{} is trained on the original dataset and the augmented dataset, respectively.}
\resizebox{\columnwidth}{!}{
\begin{tabular}{c|c|ccc|ccc}
\hline
\multirow{2}{*}{Dataset} & \multirow{2}{*}{$\delta$} & \multicolumn{3}{c|}{Group I}                        & \multicolumn{3}{c}{Group II}                         \\ \cline{3-8} 
                        &                        & $\alpha=0.25$              & $\alpha=0.5$              & $\alpha=0.75$              & $\alpha=0.25$              & $\alpha=0.5$              & $\alpha=0.75$              \\ \hline
Original   & \multirow{2}{*}{0.5}  & 0.0199 (0.0031) & 0.0182 (0.0029) & 0.0180 (0.0039) & 0.0119 (0.0015) & 0.0095 (0.0030) & 0.0112 (0.0026) \\
Augmented                    &                        & 0.0182 (0.0036) & 0.0176 (0.0040) & 0.0171 (0.0041) & 0.0129 (0.0013) & 0.0099 (0.0025) & 0.0100 (0.0026) \\ \hline
Original   & \multirow{2}{*}{2}   & 0.0370 (0.0159) & 0.0311 (0.0128) & 0.0330 (0.0126) & 0.0254 (0.0150) & 0.0148 (0.0046) & 0.0156 (0.0052) \\
Augmented                &                        & 0.0273 (0.0120) & 0.0254 (0.0117) & 0.0257 (0.0118) & 0.0218 (0.0067) & 0.0126 (0.0035) & 0.0127 (0.0037) \\ \hline
\end{tabular}}\label{tab:fpca}
\end{table}

According to the results, data augmentation leads to a substantial enhancement in predictive performance for both labels and signals when $\delta = 2$.
This implies that our data augmentation approach is useful especially when a significant signal deviation exists but lacks enough training signals. Our method that generates realistic synthetic signals provides sufficient training signals to characterize their underlying distributions in such cases. Meanwhile, \model{} trained only on the original signals gives satisfactory performances when $\delta = 0.5$. This shows that it would be enough to train only on the original dataset if the original training signals are already sufficient to characterize the distribution effectively. Finally, the accuracy of the label prediction increases as $\alpha$ increases. This is expected since signal discrepancies among different groups become more prominent as more context observations are available.



\end{document}